\documentclass{article}

\usepackage[final]{corl_2026} % Uncomment for the camera-ready ``final'' version.
\usepackage[utf8]{inputenc} % allow utf-8 input
\usepackage[T1]{fontenc}    % use 8-bit T1 fonts
\usepackage{hyperref}       % hyperlinks
\usepackage{url}            % simple URL typesetting
\usepackage{booktabs}       % professional-quality tables
\usepackage{amsfonts}       % blackboard math symbols
\usepackage{nicefrac}       % compact symbols for 1/2, etc.
\usepackage{microtype}      % microtypography
\usepackage{xcolor}         % colors
\usepackage{pifont}

\usepackage{makecell}

\usepackage{multicol}
\usepackage{subcaption}
\usepackage{mathrsfs}
\usepackage{amsfonts}
\usepackage{colortbl}               
\usepackage{amsfonts}       
\usepackage{nicefrac}       
\usepackage{microtype}  
\usepackage{marvosym}
\usepackage{array} 
\usepackage{graphicx}
\usepackage{amsmath}
\usepackage{amssymb}
\usepackage{multirow}
\setcitestyle{numbers,square}
\usepackage{algorithm}  
\usepackage{algorithmicx} 
\usepackage{algpseudocode}
\usepackage{caption}
\usepackage{subfloat}
\usepackage{tabularx} 
\usepackage{ragged2e} 
\usepackage{makecell}
\usepackage{amsfonts}
\usepackage{xspace}% blackboard math symbols
\usepackage{anyfontsize}
\usepackage{nicefrac}  
\usepackage[accsupp]{axessibility} 
\usepackage{float}
\usepackage{cleveref}
\usepackage{wrapfig}

\title{LiMA: Bridging Long-term Imagination to Real-time Dexterous Manipulation via Asynchronous  Diffusion}

\author{
        \vspace{-1.6em} \\
        \textbf{Ning Chen\textsuperscript{1,2}$^{*}$, Yankai Fu\textsuperscript{1,2}$^{*}$, 
        Junkai Zhao\textsuperscript{2}$^{\dagger}$, 
        Qianpu Sun\textsuperscript{1},} \\
        \textbf{Guocai Yao\textsuperscript{2}, 
        Pengwei Wang\textsuperscript{2}, 
        Zhongyuan Wang\textsuperscript{2}, 
        Shanghang Zhang\textsuperscript{1,2}\textsuperscript{\Letter} \vspace{0.3em}} \\
    \textsuperscript{1}State Key Laboratory of Multimedia Information Processing, School of Computer Science, \\
    Peking University; \textsuperscript{2}Beijing Academy of Artificial Intelligence \\
    $^{*}$Equal contribution, $^{\dagger}$Project leader, \textsuperscript{\Letter}Corresponding author
    \vspace{0.3em}\\
    \textbf{Project Webpage:} \href{https://ccdcs.github.io/LiMA_repo/}{LiMA\ Project}
    \vspace{-0.9em}
}

\begin{document}
\maketitle

%===============================================================================

\begin{center}
    \vspace{-1em}
    \captionsetup{type=figure} % 确保在非figure环境下也能正常显示caption
    % trim={左 下 右 上}
    \includegraphics[width=0.95\textwidth]{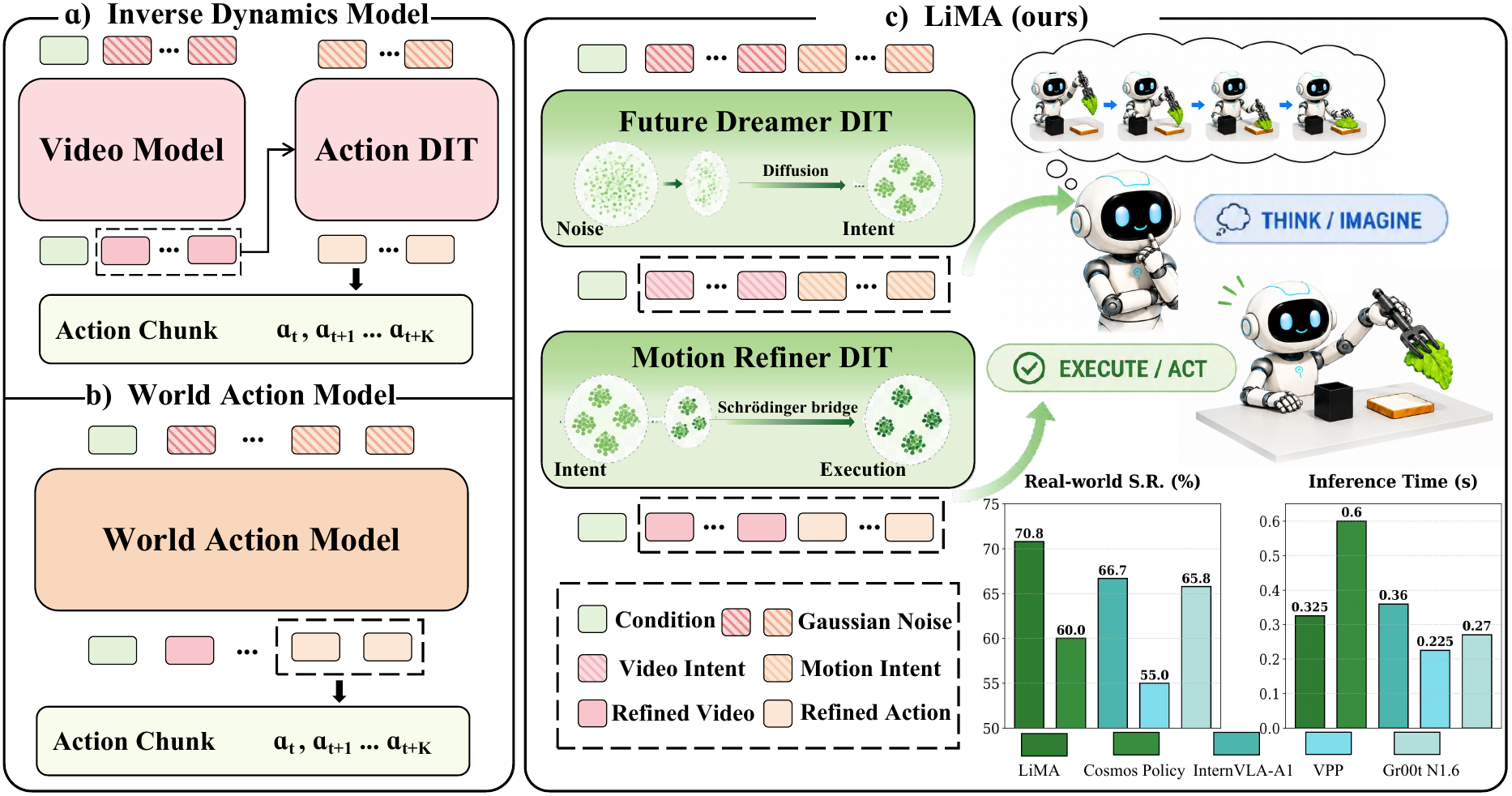}
    \captionof{figure}{
Unlike (a) Inverse Dynamics which sequentially generate observations before predicting actions, or (b) World Action Models that jointly denoise images and actions, we propose (c) LiMA, an asynchronous hierarchical framework. By decoupling long-term imagination from high-frequency execution, LiMA bridges latent foresight with multi-scale refinement, ensuring both strategic planning and real-time responsiveness across diverse horizons.}
    \label{fig:Teaser}
\end{center}
\vspace{-0.8em}

\begin{abstract}
Dexterous manipulation demands long-term foresight and rapid reactive control. Vision-Language-Action (VLA) models, while proficient in high-level reasoning, often lack a fine-grained understanding of physical dynamics and spatial perception. Conversely, World-Action Models (WAMs) typically suffer from high inference latency due to iterative generation. These deficiencies result in a critical temporal misalignment where the model's intent fails to adapt to rapid physical contact changes. To overcome this fundamental bottleneck, we propose \textbf{LiMA}, an asynchronous dual-system generative framework that systematically decouples intent planning from reactive execution. LiMA organizes computation into a multi-scale hierarchy: a slow system handles sparse long-horizon spatiotemporal intent generation, while a fast system focuses on dense high-frequency motion refinement. To align sparse intent predictions with dense action trajectories, we introduce a Latent Schrödinger Bridge Coupling mechanism that formulates refinement as an entropy-regularized probabilistic transport process. LiMA reduces inference latency by \textbf{45.8\%} compared with Cosmos-Policy via asynchronous decoupling. Evaluated across six bimanual dexterous manipulation tasks spanning multiple horizons, LiMA achieves an overall success rate of \textbf{70.8\%} and an average subtask success rate of \textbf{78.9\%}, while maintaining performance in unseen scenarios.
\end{abstract}


%===============================================================================

\section{Introduction}
Dexterous manipulation represents one of the most sophisticated challenges in robotics, requiring a seamless integration of long-term temporal foresight and rapid reactive precision~\cite{fu2025cordvip, dexvlg25, bai2025towards}. Recent Vision-Language-Action (VLA) models have made substantial progress by scaling pretraining on large egocentric manipulation datasets~\cite{punamiya2026egoverse, chen2026action100m}, enabling them to acquire broad visual, linguistic, and action priors for dexterous tasks~\cite{yang2025egovla, luo2026being, hoque2025egodex, fu2025metis, cai2025n, team2024octo, zitkovich2023rt}. Despite these advancements, existing VLA models still lack an explicit mechanism for anticipating and responding to the continuous evolution of the physical world, which is crucial for robust and generalizable dexterous manipulation.

World-Action Models (WAMs) have emerged as a promising paradigm, leveraging pretrained video generative models~\cite{chi2025wow, goswami2026dexwm} to internalize pixel-level physical world knowledge and exhibiting superior generalization and generative diversity. Current WAM frameworks generally follow two paths: (1) Inverse Dynamics Models (IDMs)~\cite{jia2025video2actdualsystemvideodiffusion, hu2024video} employ a video model to "imagine" future observations as a conditional prior for action generation, improving policy generalization through future-aware visual priors. However, this sequential conditioning design loosely couples visual imagination with action prediction, limiting cross-modal feature interaction. (2) Integrated WAMs~\cite{kim2026cosmos, lingbot-va2026, ye2026worldactionmodelszeroshot, cen2025rynnvla} perform joint image-action generation within a unified denoising process, enabling richer spatiotemporal interactions. Yet, they are hindered by an excessively long latent feature space that leads to prohibitive inference latency, failing to meet the high-frequency reactive requirements of dexterous manipulation.

In this work, we introduce \textbf{LiMA}, an asynchronous multi-scale framework for jointly generating long-term foresight and robotic actions. LiMA employs a dual-system architecture that coordinates generation across distinct temporal scales, enabling rapid physical execution without sacrificing long-horizon planning. To maintain coherence between the two systems, we propose a Latent Schrödinger Bridge Coupling mechanism that models the transition from abstract intent to motor commands as a continuous probabilistic transport process, tightening the feature coupling between high-level imagination and low-level refinement. This integration is further enhanced by Spatiotemporal Adaptive Modulation, which grounds long-term foresight into real-time physical observations.

To evaluate LiMA, we conduct extensive experiments on diverse dexterous manipulation tasks across varying temporal horizons. Comparative analysis demonstrates that LiMA achieves a stronger balance between task success and execution efficiency. Compared with state-of-the-art VLA~\cite{gr00tn1_2025, internvla_a1} and WAM~\cite{hu2024video, kim2026cosmos} baselines, LiMA improves success rates especially in long-horizon and contact-intensive tasks, where stable task-level intent and fine-grained reactive correction are both required. Compared with integrated WAM baselines~\cite{kim2026cosmos}, LiMA substantially reduces inference latency through asynchronous coordination between long-range imagination and high-frequency action refinement, while maintaining their latent-level coupling. In summary, our contributions are as follows:
\begin{itemize}
    \item We present \textbf{LiMA}, an asynchronous dual-system framework that decouples long-range imagination from reactive execution, overcoming generative inference bottlenecks.
    \item We introduce a Latent Schrödinger Bridge mechanism with adaptive modulation to ensure tight feature alignment between semantic intent and physical action.
    \item We provide a thorough evaluation demonstrating that LiMA achieves competitive results on real-world tasks while offering superior computational efficiency for real-time applications.
\end{itemize}
% \end{itemize}

%===============================================================================
\section{Related Work}
\subsection{Dexterous Manipulation}
Dexterous manipulation poses a pioneering challenge due to high-dimensional action spaces and complex contact dynamics~\cite{zhang2026fingervip, li2026deco, qi2023general, funabashi2022multi, ye2026datapyramid}. Early point-cloud-based methods~\cite{ze20243d, fu2025cordvip, zhong2025dexgrasp} handled contact physics but lacked semantic depth. To address this, recent Vision-Language-Action (VLA) models~\cite{Zawalski24-ecot, intelligence2025pi_, kim2024openvla, lin2025onetwovlaunifiedvisionlanguageactionmodel, ye2025token} distilled large-scale reasoning priors into control, yet they often fail to ground abstract intent into fine-grained physical interactions. To better capture environment physics, World-Action Models (WAMs)~\cite{pai2025mimicvideo, bi2025motus, li2026world, yuan2026fastwam, harmowam2026, lou2026dreamtac} employ video synthesis as a generative prior, leveraging pixel-level dynamics to internalize robust physical world knowledge and improve out-of-distribution generalization. However, existing WAMs face a critical trade-off: sequential Inverse Dynamics Models (IDMs) suffer from sparse feature coupling, while unified generative models are bottlenecked by the computational cost of joint denoising. In contrast, LiMA introduces an asynchronous hierarchical framework that leverages video-based foresight while maintaining high-frequency reactive execution, reconciling physical grounding with real-time responsiveness.

\subsection{Hierarchical Planning and Control in Robotics}
To harmonize long-horizon reasoning with real-time sensorimotor control, recent literature has shifted toward dual-system hierarchies, loosely inspired by the cognitive division between deliberation and reaction~\cite{xue2025reactive, xu2025diffusionbasedimaginativecoordinationbimanual, song2025hume}. Initial explorations focused on synchronous coordination, where a high-level cognitive model (System 2) generates latent embeddings to condition a lower-level policy head (System 1)~\cite{zhao2025cotvla, bu2025univla, song2025rationalvla, wen2025dexvla}. Despite their structural clarity, these methods are often bottlenecked by the lower sampling frequency of the deliberative module, failing to exploit the potential of System 1 for agile, high-rate execution. To break this temporal constraint, asynchronous architectures have emerged~\cite{liu2026last, chen2025fastinslowdualsystemfoundationmodel, galaxea2025, zhang2024hirt}, decoupling the refresh rates of task-level foresight and action-level response. However, current asynchronous designs treat the output of System 2 as a static extrinsic condition. This superficial coupling leads to fragile alignment and poor inheritance of reasoning priors. LiMA bridges this gap by interlinking both systems at the noise level via a distributional diffusion bridge. Instead of simple concatenation, our approach anchors the reactive controller to the generative manifold of System 2, ensuring multimodal consistency with real-time agility.

\section{Method}

\subsection{Preliminaries}
\label{Preliminaries}
We formulate the sequential decision-making process of bimanual dexterous manipulation as a conditional generative problem. At each execution time step $t$, the system receives multi-modal sensory inputs comprising a language task instruction $L$, the robot's current proprioceptive joint state $S_t$, and multi-view visual observations consisting of three camera streams: a head-mounted ego-centric view and two wrist-mounted views, denoted as $\mathbf{O}_t = \{I_{t}^{\text{ego}}, I_{t}^{\text{left}}, I_{t}^{\text{right}}\}$. 

The global objective of the generative policy is to model the conditional joint probability distribution to predict a synchronized action chunk $\mathcal{A}$ over a future temporal horizon, which encapsulates the coordinated trajectories of both the robotic arms and all dexterous fingers:
\begin{equation}
\mathcal{A}_{t:t+K} \sim p\left( \mathcal{A} \mid \mathbf{O}_t, S_t, L \right)
\end{equation}

To parameterize this inter-system transition, we adopt the Schrödinger Bridge framework~\cite{liu2023i2sb}, which constructs an entropy-regularized optimal transport path between two distributions. Unlike standard diffusion generation, which typically starts from an unstructured Gaussian prior, the Schrödinger Bridge provides a structured boundary-to-boundary formulation. In our setting, this allows the reactive controller to start from the Dreamer's strategic intent prior $X_1$ and progressively refine it toward the Refiner's clean execution latent $X_0$. Specifically, the intermediate state $X_t$ is sampled from a conditional Gaussian posterior $q(X_t \mid X_0, X_1)$ along a deterministic interpolation path. This formulation enables a direct bridge-matching objective to learn the transition from the Dreamer's intent prior to the Refiner's execution latent, rather than relying on an unconditional noise prior. The detailed derivation and variance scheduling are provided in the Appendix.

\begin{figure}
    \centering
    \includegraphics[width=\textwidth]{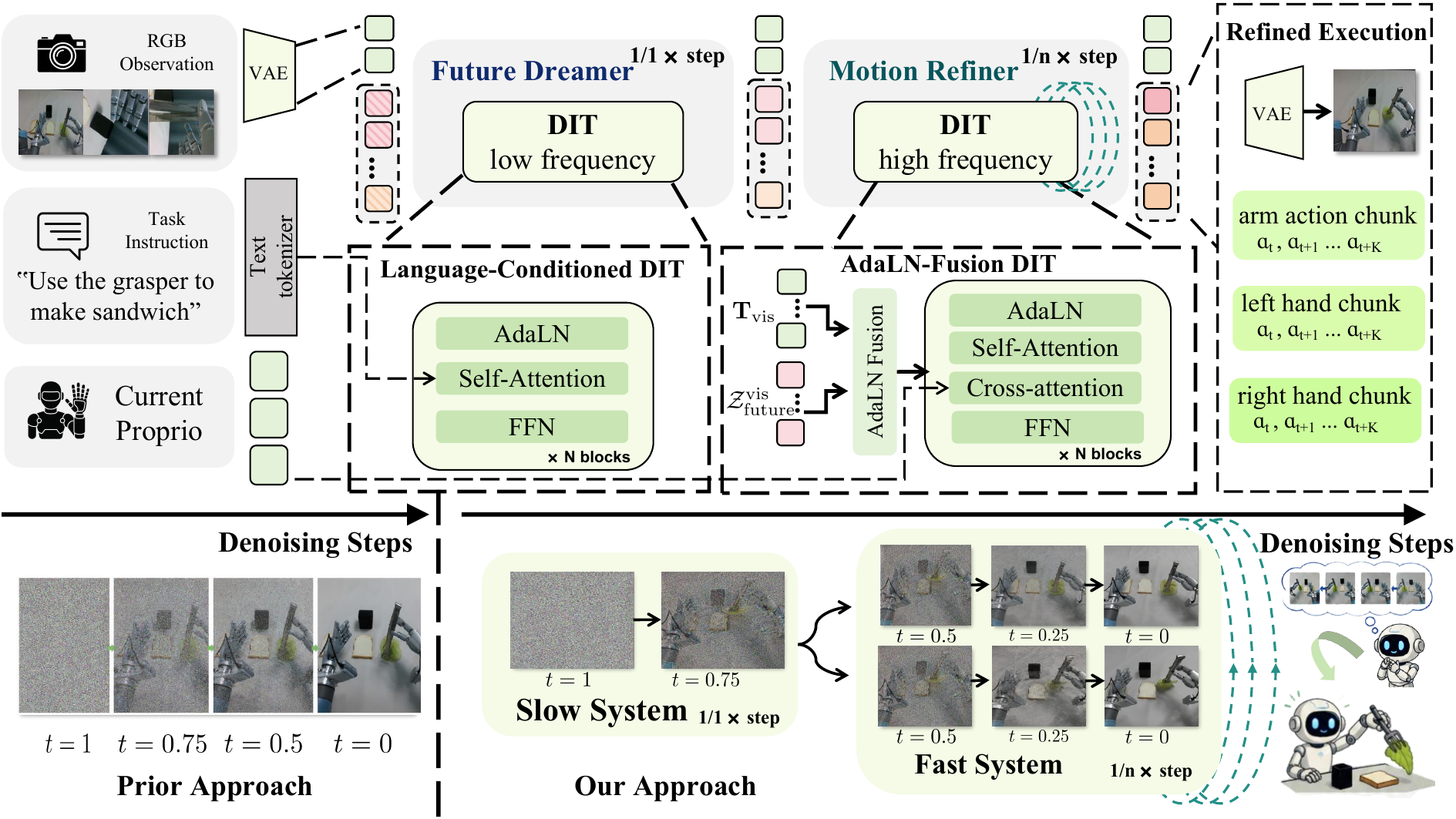}
    \caption{\textbf{Overview of LiMA.} LiMA asynchronously coordinates long-horizon foresight and real-time reactive control through a dual-system architecture. The Future Dreamer produces structured intent, while the Motion Refiner performs high-frequency latent refinement. A Latent Schrödinger Bridge Coupling mechanism maintains latent-level coherence between the two systems for manipulation.}
    \label{fig:pipeline}
    \vspace{-2em}
\end{figure}

\subsection{Model Architecture}
\label{Model Architecture}
LiMA adopts a dual-DiT~\cite{Peebles2022DiT} architecture to integrate high-level generative imagination with low-level reactive control. As illustrated in Figure~\ref{fig:pipeline}, the framework hierarchically channels the preliminary denoising output of a Future Dreamer—which synthesizes a long-term spatiotemporal task imagination—as an informed latent prior into a Motion Refiner. This enables LiMA to effectively couple global strategic foresight with real-time action execution within a unified latent feature space.

\noindent\textbf{Unified Multimodal Representation.} LiMA maps heterogeneous sensory inputs into a unified latent sequence within a shared embedding space $\mathbb{R}^D$. Specifically, we employ a pre-trained Wan2.1 spatiotemporal VAE tokenizer~\cite{wan2025} to compress multi-view visual frames $\mathbf{O}_t$ into visual tokens $\mathbf{T}_{\text{vis}}$, and a frozen T5-XXL encoder~\cite{2020t5} to extract language embeddings $\mathbf{L}$. Crucially, the robot's proprioceptive state $\mathbf{S}_t$ is directly unified into the visual feature dimension via a structural repetitive padding mechanism to form state tokens $\mathbf{T}_{\text{state}}$, circumventing parametric projection layers.

\noindent\textbf{Future Dreamer Flow (Slow System).} 
The Future Dreamer ($\mathcal{G}_{\text{dream}}$) models long-horizon spatiotemporal task evolution as the slow branch of LiMA. It is initialized from a pre-trained Cosmos-Predict2-2B~\cite{nvidia2025cosmosworldfoundationmodel} to inherit physical world priors, and is further fine-tuned on task-specific demonstrations. The Dreamer adopts an asymmetric conditioning design: visual tokens $\mathbf{T}_{\text{vis}}$ form the main denoising stream, while language embeddings $\mathbf{L}$ modulate Transformer activations through Adaptive Layer Normalization (AdaLN)~\cite{perez2018film}. After preliminary denoising, it outputs a long-horizon intent representation $\mathcal{Z}_{\text{int}} = [\mathcal{Z}_{\text{future}}^{\text{vis}}, \mathcal{Z}_{\text{future}}^{\text{act}}]$, which contains imagined multi-view visual latents and a coarse macro-action prior. This intent representation is organized to match the Refiner's latent interface, allowing it to serve as the structured prior boundary for bridge refinement.

\noindent\textbf{Motion Refiner Flow (Fast System).} 
The Motion Refiner ($\mathcal{G}_{\text{refine}}$) acts as the fast branch for high-frequency action refinement. Unlike the pre-trained Dreamer, $\mathcal{G}_{\text{refine}}$ is a smaller 1B-parameter Transformer trained from scratch, enabling it to specialize in fine-grained motor coordination. Given the intent prior $\mathcal{Z}_{\text{int}}$, the Refiner uses a decoupled conditioning scheme that separately handles imagined visual latents, coarse action priors, and real-time proprioceptive states. It predicts a joint execution latent $\mathcal{Z}_{\text{out}} = [\mathcal{Z}_{\text{out}}^{\text{act}}, \mathcal{Z}_{\text{out}}^{\text{vis}}]$, which contains the near-term action chunk and the short-horizon ego-centric visual latent. The coarse action prior $\mathcal{Z}_{\text{future}}^{\text{act}}$ is replicated along the action horizon to initialize the action refinement trajectory, while the imagined visual prior $\mathcal{Z}_{\text{future}}^{\text{vis}}$ is fused with online visual tokens through a localized adaptive modulation layer, as shown in Figure~\ref{fig:pipeline}.

Specifically, we exploit the spatial correspondence between imagined future views and streaming observations. For each camera slot $i$, the corresponding online visual tokens $\mathbf{T}_{\text{vis}}^{(i)}$ and imagined visual prior $\mathcal{Z}_{\text{future}}^{\text{vis},(i)}$ are integrated through adaptive modulation:
\begin{equation}
\label{eq:adaln_fusion_pure}
\mathcal{T}_{\text{mod}}^{(i)} = 
\gamma(\mathcal{Z}_{\text{future}}^{\text{vis},(i)}) 
\cdot 
\text{LayerNorm}(\mathbf{T}_{\text{vis}}^{(i)}) 
+ 
\beta(\mathcal{Z}_{\text{future}}^{\text{vis},(i)}),
\end{equation}
where $\gamma$ and $\beta$ are affine scale and shift parameters predicted from the imagined visual prior. This modulation allows the Refiner to inject long-horizon visual intent into real-time observations while preserving sensitivity to local scene changes. In parallel, the instantaneous state tokens $\mathbf{T}_{\text{state}}$ are projected into continuous embedding streams and used as proprioceptive conditioning signals throughout the Refiner, keeping motion refinement grounded in the latest robot state.

\subsection{Latent Schrödinger Bridge Coupling}
\label{sec:latent_schrodinger_bridge}
Building on the I2SB formulation detailed in the Appendix, we construct a latent bridge between the Dreamer and the Refiner within a shared latent space. Instead of starting action refinement from an unconditional Gaussian prior, we use the Future Dreamer's denoised intent representation $\mathcal{Z}_{\text{int}}$ as the structured prior boundary, corresponding to $X_1$, and the ground-truth execution latent $\mathcal{Z}_{\text{out}}^{(0)}$ as the clean data boundary, corresponding to $X_0$. Since $\mathcal{Z}_{\text{int}}$ is organized to match the Refiner's latent interface, the bridge can be constructed directly without an additional alignment projection.

Given this boundary pair, the intermediate reactive bridge latent $\mathcal{Z}_{\text{out}}^{(t)}$ is sampled from the conditional posterior:
\begin{equation}
\label{eq:lima_posterior}
\mathcal{Z}_{\text{out}}^{(t)} \sim q\left(\mathcal{Z}_{\text{out}}^{(t)} \mid \mathcal{Z}_{\text{out}}^{(0)}, \mathcal{Z}_{\text{int}}\right) 
= \mathcal{N}\left( 
\mathcal{Z}_{\text{out}}^{(t)}; \,
\mu_t\left(\mathcal{Z}_{\text{out}}^{(0)}, \mathcal{Z}_{\text{int}}\right), \,
\Sigma_t \mathbf{I} 
\right),
\end{equation}
where the mean path $\mu_t$ defines the interpolation between the clean execution latent and the intent-conditioned prior, while $\Sigma_t$ controls the stochastic perturbation along the bridge.

The Motion Refiner is trained to directly predict the clean execution latent from the intent-conditioned bridge state:
\begin{equation}
\label{eq:refine_loss}
\mathcal{L}_{\text{refine}} =
\mathbb{E}_{t, \mathcal{Z}_{\text{out}}^{(0)}, \mathcal{Z}_{\text{int}}}
\left[
\left\|
\hat{\mathcal{Z}}_{\text{out}}^{(0)}
-
\mathcal{Z}_{\text{out}}^{(0)}
\right\|_2^2
\right],
\quad
\hat{\mathcal{Z}}_{\text{out}}^{(0)}
=
\mathcal{G}_{\theta}
\left(
\mathcal{Z}_{\text{out}}^{(t)}, t, \mathcal{Z}_{\text{int}}
\right).
\end{equation}
This objective encourages the Refiner to recover the clean joint execution latent from an intent-conditioned intermediate state, rather than denoising from an unconditional Gaussian prior.

During real-time inference, the clean boundary $\mathcal{Z}_{\text{out}}^{(0)}$ is unavailable. Therefore, we initialize the refinement process from the intent-conditioned prior:
\begin{equation}
\mathcal{Z}_{\text{out}}^{(1)} \sim 
\mathcal{N}\left(
\mathcal{Z}_{\text{int}}, 
\sigma_1^2 \mathbf{I}
\right).
\end{equation}
Starting from this structured prior reduces the sampling burden of the Refiner and keeps the generated action trajectory aligned with the Dreamer's long-horizon intent. As a result, the fast system can focus on local motion correction while preserving consistency with the high-level foresight.

\subsection{Asynchronous Dual-System Coordination}
\label{Asynchronous Dual-System Coordination}
To deploy compute-intensive generative loops for real-time control, LiMA adopts an asynchronous temporal decoupling mechanism to resolve the inherent trade-off between cognitive capacity and inference latency. Specifically, while the Future Dreamer performs long-horizon spatiotemporal imagination to provide global strategic foresight, its heavy computational cost limits execution frequency. To bridge this mismatch, we introduce a periodic execution interval $\Delta T \in \mathbb{Z}^+$ to decouple the inference frequencies of the dual systems.

Formally, at update steps where $t \bmod \Delta T = 0$, the compute-heavy Future Dreamer is triggered to process the current observation $\mathbf{O}_t$ and language instruction $L$, generating a refreshed strategic intent prior $\mathcal{Z}_{\text{int}, t}$ in the unified latent space. The Motion Refiner simultaneously consumes this updated prior together with $\mathbf{O}_t$ to produce the fine-grained execution chunk $\mathcal{Z}_{\text{out}, t}$.

At intermediate non-update steps where $t \bmod \Delta T \neq 0$, the Future Dreamer bypasses inference and freezes its latent output. Meanwhile, the lightweight Motion Refiner remains active at full operational frequency, continuously integrating the latest observation $\mathbf{O}_t$ with the cached strategic prior $\mathcal{Z}_{\text{int}, \lfloor t/\Delta T \rfloor \Delta T}$ to synthesize real-time actions. This sparse planning coupled with dense, high-frequency motion refinement enables robust reactivity under unforeseen environmental perturbations.

\begin{table}
    \renewcommand{\arraystretch}{1}
    \centering
    \caption{\textbf{Comparison of LiMA and baselines on Real-World Tasks.} Each experiment is evaluated with 20 trials. Inference speed is evaluated on a NVIDIA H100 GPU.}
    \label{tab:Main_Experiments}
    \resizebox{\linewidth}{!}{
    \begin{tabular}{l | cc cc cc cc cc cc | c} % 1 + 12 + 2 + 1 列定义
        \toprule
        % 使用 multicolumn{1}{c} 确保 Method 左右没有竖线穿透顶部
        \multicolumn{1}{c}{\bf Method} & 
        \multicolumn{2}{c}{\bf Stack cup} & 
        \multicolumn{2}{c}{\bf Roll T-shirt} &
        \multicolumn{2}{c}{\bf Cook Rice} &  
        \multicolumn{2}{c}{\bf Make Sandwich} & 
        \multicolumn{2}{c}{\bf Make Coffee} & 
        \multicolumn{2}{c}{\bf Assemble package} &
        \multicolumn{1}{c}{\bf Time (ms)} \\ 
        
        \cmidrule(lr){2-13} \cmidrule(lr){14-14}
        
        & \bf SR & \bf PSR & \bf SR & \bf PSR & \bf SR & \bf PSR & \bf SR & \bf PSR & \bf SR & \bf PSR & \bf SR & \bf PSR & \bf - \\
        \midrule
        GR00T N1.6    & 85.0\% & 87.5\%  & 70.0\% & 71.7\% & 70.0\% & 72.0\%  & 65.0\% & 72.5\% & 55.0\% & \textbf{71.7}\% & \textbf{50.0\%} & 61.7\% & 270 \\ 
        VPP        & 80.0\% & 80.0\% & 55.0\% & 56.7\% & 55.0\% & 60.0\%  & 50.0\% & 61.3\% & 50.0\% & 56.7\% & 40.0\% & 48.3\% & \textbf{225} \\
        Internvla-a1  & \textbf{90.0\%} & 90.0\% & \textbf{75.0\%} & \textbf{88.3\%} & 70.0\% & 78.0\%  & 65.0\% & 77.5\% & 50.0\% & 66.7\% & \textbf{50.0\%} & 58.3\% & 360 \\
        cosmos-policy & 80.0\% & 82.5\% & 60.0\% & 70.0\% & 60.0\% & 67.0\%  & 60.0\% & 71.3\% & 55.0\% & 63.3\% & 45.0\% & 51.7\% & 600 \\
        \rowcolor{gray!20} \bf LiMA (Ours) & \textbf{90.0\%} & \textbf{95.0\%} & \textbf{75.0\%} & 83.3\% & \textbf{80.0\%} & \textbf{82.0\%} & \textbf{70.0\%} & \textbf{80.0\% }& \textbf{60.0\%} & 70.0\% & \textbf{50.0\%} & \textbf{63.3\%} & 325 \\
        \bottomrule
    \end{tabular}
}
    % \vspace{-1.2em}
\end{table}

\begin{figure}
    \centering
    \begin{minipage}[t]{0.41\linewidth}
        \centering
        \includegraphics[width=\linewidth]{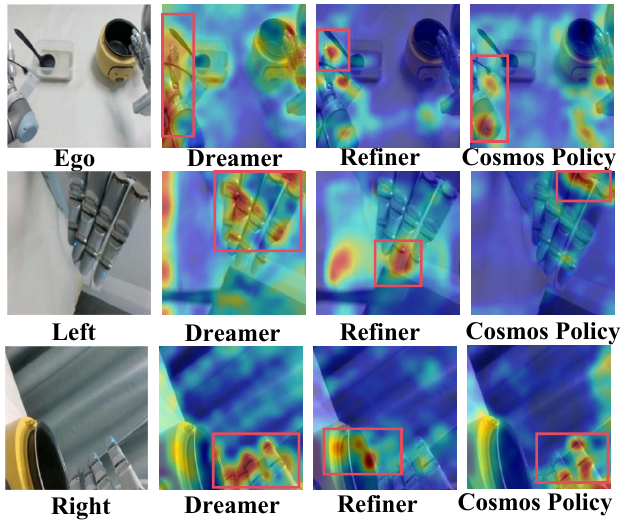}
        \caption{\small \textbf{Attention Map Comparison.}}
        \label{fig:Attention_Map}
        % \vspace{-1.5em}
    \end{minipage}
    \hfill
    \begin{minipage}[t]{0.57\linewidth}
        \centering
        \includegraphics[width=\linewidth]{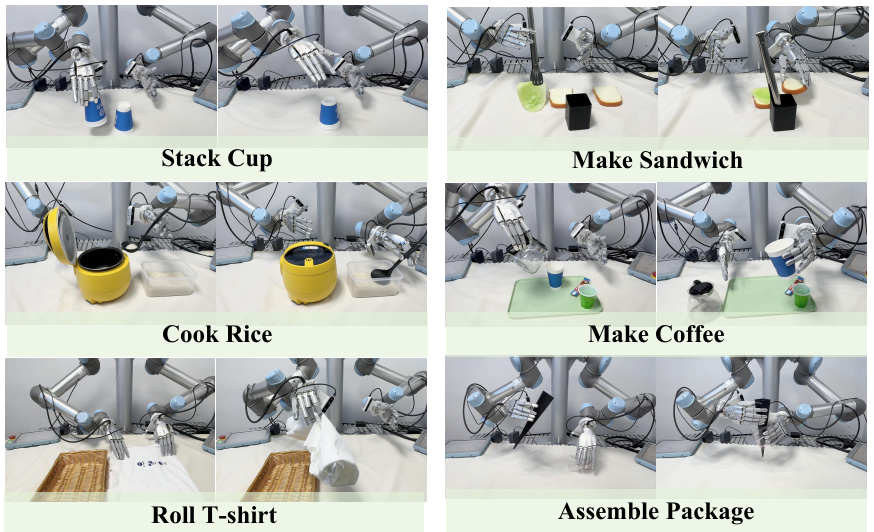}
        \caption{\textbf{Visualization of Dexterous Tasks.}}
        \label{fig:tasks}
    \end{minipage}
    
    \vspace{-1em}

\end{figure}

\subsection{Training Objective}
\label{sec:training_objective}
We train LiMA with a joint optimization objective over the Future Dreamer and the Motion Refiner. For the Future Dreamer, the learning objective $\mathcal{L}_{\text{dream}}$ is a joint-space score-matching loss that supervises the generation of long-horizon action chunks and three-view future visual latents. To improve the robustness of the Dreamer--Refiner interface during real-time inference, we introduce a Boundary Noise Injection regularizer on the conditioning path of the fast system. Specifically, during training, the Refiner receives a lightly perturbed copy of the intent latent with a small probability $p_{\text{bni}}$:
\begin{equation}
\tilde{\mathcal{Z}}_{\text{int}} =
\begin{cases}
\mathcal{Z}_{\text{int}} + \gamma \epsilon_{\text{noise}}, 
& \text{with probability } p_{\text{bni}}, \\
\mathcal{Z}_{\text{int}}, 
& \text{otherwise},
\end{cases}
\quad
\epsilon_{\text{noise}} \sim \mathcal{N}(0,\mathbf{I}),
\quad
\gamma \ll 1 .
\end{equation}
Conditioned on $\tilde{\mathcal{Z}}_{\text{int}}$, the Motion Refiner is optimized with $\mathcal{L}_{\text{refine}}$ to directly recover the clean joint execution latent $\mathcal{Z}_{\text{out}}^{(0)}$ from an intent-conditioned bridge state. This target contains fine-grained dual-arm actions, left/right anthropomorphic hand motions, and near-term ego-centric visual latents. The total training objective is:
\begin{equation}
\mathcal{L}_{\text{total}} =
\mathcal{L}_{\text{dream}} + \lambda \mathcal{L}_{\text{refine}} .
\end{equation}

\section{Experiment}
\label{Experimental}

\subsection{Experiment Setup}
\label{Experiment Setup}
\noindent\textbf{Robot Setup.} Our real-world evaluation platform consists of a bimanual setup featuring two 6-DoF UR5 robotic arms equipped with two 22-DoF SharpaWave five-fingered dexterous hands. The multimodal perception suite comprises three Intel RealSense D435 cameras: one head-mounted camera capturing the global ego-centric view and two wrist-mounted cameras capturing localized coordination views. Following~\cite{fu2025metis}, we use VIVE Trackers to capture the relative spatial trajectories of the human wrists for real-time end-effector control of the UR5 arms. Concurrently, human finger configurations are recorded via MetaGlove Pro gloves and mapped onto the 22-DoF SharpaWave hands through a joint-space kinematic retargeting pipeline.

\noindent\textbf{Tasks Setup.} We evaluate LiMA across two short-horizon tasks—\textit{Stack Cup}, \textit{Roll T-shirt}—and four challenging long-horizon bimanual dexterous tasks, as shown in \Cref{fig:tasks}:  \textit{Cook Rice}, \textit{Make Sandwich}, \textit{Make Coffee}, and \textit{Assemble Package}. For each task, we collect 100 high-quality expert teleoperation demonstrations. During physical deployment, the policy is evaluated in 20 trials per task, and the success rate is reported as the main metric.

\noindent\textbf{Baselines and Evaluation Protocol.}We evaluate LiMA against four state-of-the-art baselines, including two Vision-Language-Action (VLA) models — Gr00T N1.6~\cite{gr00tn1_2025} and InternVLA-a1~\cite{internvla_a1} — and two World-Action Models (WAM) —VPP~\cite{hu2024video} and Cosmos-Policy~\cite{kim2026cosmos}. To rigorously quantify performance across diverse dexterous scenarios, we employ Success Rate (SR) to measure overall task completion and Progress Success Rate (PSR) as a fine-grained metric for subtask advancement.
%===============================================================================

\subsection{Results and Analysis}
\noindent\textbf{Real-World Quantitative Analysis.} 
LiMA achieves strong overall performance across the six real-world manipulation tasks, as shown in \Cref{tab:Main_Experiments}. On short-horizon tasks such as Stack Cup and Roll T-shirt, LiMA performs comparably to strong VLA baselines, indicating that the hierarchical generative design does not compromise basic manipulation accuracy. More importantly, on long-horizon and contact-intensive tasks such as Cook Rice and Make Sandwich, LiMA shows clearer advantages in task progress and execution stability, as reflected by its improved PSR and competitive SR. This suggests that the Dreamer's long-horizon intent helps the Refiner maintain task progress and reduce intermediate failures during complex bimanual interactions. Compared with unified WAM baselines, LiMA further improves the efficiency--performance trade-off, reducing inference latency from 600ms for Cosmos-Policy to 325ms, corresponding to a 45.8\% latency reduction. These results show that LiMA maintains strong task performance while enabling more efficient real-time execution for high-frequency closed-loop dexterous manipulation.

\noindent\textbf{Attention-Map Analysis.} As shown in \Cref{fig:Attention_Map}, we visualize and compare the attention maps of action tokens across three camera views for Cosmos-Policy, which adopts a single DiT architecture, and LiMA, which consists of the Dreamer and the Refiner. In the Cook Rice task, the robot is required to grasp a ladle with the left hand while keeping the right hand relatively stationary, making both local manipulation accuracy and bimanual coordination important. We observe that Cosmos-Policy produces relatively dispersed attention, with noticeable weights assigned to regions that are less relevant to the current manipulation. In contrast, LiMA exhibits a clearer hierarchical division of attention. The Dreamer attends more to global spatial relationships and task-relevant context, supporting long-horizon intent generation, while the Refiner focuses more on localized manipulation regions for fine-grained grasping and real-time motion adjustment. These visualizations provide qualitative evidence that LiMA can separate long-term planning from local execution, leading to more stable interaction in complex bimanual manipulation tasks.

\noindent\textbf{Latency Analysis.}
We measure end-to-end inference latency from receiving the multi-view observations to decoding an executable action chunk. Under the asynchronous execution schedule, the Dreamer is refreshed once every four Refiner updates ($\Delta T=4$). We therefore report the amortized latency per Refiner update, computed as
$T_{\mathrm{avg}}=(T_{\mathrm{Dreamer}}+4T_{\mathrm{Refiner}})/4$.
LiMA requires 325\,ms on average to produce a 32-step action chunk. Therefore, the reported value represents chunk-generation latency rather than latency per individual control action. For a fair comparison, all methods are evaluated on the same NVIDIA H100 GPU using the same batch size and numerical precision, without model-specific operator fusion, model compilation, quantization, or edge-device deployment optimization.

\subsection{Ablation Study}
\noindent\textbf{Bridge Design Ablation.}
To isolate the contribution of the proposed I$^2$SB coupling, we compare LiMA with three controlled variants in \Cref{tab:bridge_ablation}: (i) removing future visual intent, (ii) initializing the Refiner from Gaussian noise while injecting the Dreamer latent through cross-attention, and (iii) replacing I$^2$SB with Flow Matching under the same architecture and conditioning setup. Removing future visual intent leads to the largest performance degradation on both tasks, confirming that the imagined visual latent provides essential task-level guidance. The Gaussian-initialized cross-attention variant also underperforms LiMA. In this variant, the Dreamer latent is supplied only as an auxiliary condition, while the Refiner must construct the execution trajectory from an unstructured Gaussian source. Such indirect conditioning may be insufficient when the few-step Dreamer prediction remains coarse or imperfect. In contrast, I$^2$SB treats the Dreamer intent and target execution latent as paired boundary states and constructs a stochastic diffusion bridge with analytically tractable intermediate marginals. This formulation directly anchors the entire refinement trajectory to the structured intent prior, rather than relying on the network to inject intent indirectly through cross-attention. Although Flow Matching also transports probability mass between distributions, our standard Flow Matching variant learns a deterministic velocity field along a prescribed probability path. The stronger performance of I$^2$SB suggests that its stochastic, boundary-anchored formulation better accommodates the uncertainty and residual mismatch between coarse Dreamer intents and fine-grained Refiner executions.

\begin{table}[t]
    \centering
    \caption{\textbf{Bridge design ablation.}
    Success rates (\%) are reported as mean $\pm$ standard deviation over three seeds, with 20 real-world trials per seed.}
    \label{tab:bridge_ablation}
    \small
    \setlength{\tabcolsep}{6pt}
    \begin{tabular}{lcc}
        \toprule
        \textbf{Variant} &
        \textbf{Cook Rice} &
        \textbf{Stack Cup} \\
        \midrule
        w/o Future Visual Intent
        & $61.7 \pm 7.6$
        & $76.7 \pm 2.9$ \\

        Gaussian Init. + Cross-Attn
        & $70.0 \pm 5.0$
        & $78.3 \pm 2.9$ \\

        Flow Matching
        & $73.3 \pm 2.9$
        & $83.3 \pm 5.8$ \\

        \textbf{LiMA}
        & $\mathbf{78.3 \pm 2.9}$
        & $\mathbf{93.3 \pm 5.8}$ \\
        \bottomrule
    \end{tabular}
\end{table}

\noindent\textbf{Component Ablation.}
To validate the effectiveness of each design choice, we first ablate LiMA's major components, as shown in \Cref{tab:comprehensive_ablation}. Removing spatiotemporal adaptive modulation decreases SR from 80\% to 60\%, indicating that simple conditioning is insufficient to align the Dreamer's long-horizon intent with online visual observations. Removing multi-action tokens further reduces SR to 55\%, and the policy tends to produce jittery and unstable motions during fine-grained grasping. This suggests that explicit action-token decomposition is important for dexterous bimanual coordination. In contrast, removing the fast--slow asynchronous mechanism does not improve task success, but increases inference latency from 0.325s to 0.45s. This shows that the asynchronous dual-system design mainly improves execution efficiency while preserving manipulation accuracy.

\noindent\textbf{Bridge Denoising Steps.}
We further analyze the influence of bridge denoising steps. When only 5 denoising steps are used, the generated intent prior remains relatively coarse, leading to a clear drop in success rate. Increasing the number of steps to 15 slightly improves the success rate, but also introduces substantially higher inference latency. Therefore, 10 denoising steps provide a more practical performance--efficiency trade-off for real-time control.

\begin{table}[t]
\centering
\caption{\textbf{Ablation Studies.} We evaluate LiMA across three dimensions: architecture components, denoising steps, and system ratios ($N_{slow}:N_{fast}$). S.R.: Success Rate, Time: Inference Latency.}
\label{tab:comprehensive_ablation}
\resizebox{\linewidth}{!}{
\begin{tabular}{l cc | l cc | l cc}
\toprule
\multicolumn{3}{c|}{\textbf{(a) Component Ablation}} & \multicolumn{3}{c|}{\textbf{(b) Denoising Steps}} & \multicolumn{3}{c}{\textbf{(c) System Ratio ($N_{s}:N_{f}$)}} \\
Component & S.R. & Time & Steps & S.R. & Time & Ratio & S.R. & Time  \\
\midrule
Full (LiMA) & \textbf{80\%} & \textbf{0.325s} & 5  & 65\% & 0.23s & 1:9 & 50\% & \textbf{0.28s} \\
w/o Adapt.  & 60\% & -   & 10 & 80\% & \textbf{0.325s} & 2:8 & 65\% & 0.31s \\
w/o Tokens  & 55\% & -     & 15 & \textbf{85\%} & 0.43s & 3:7 & \textbf{80\%} & 0.325s \\
w/o Fast-Slow     & 80\% & 0.45s & 20 & 80\% & 0.51s & 4:6 & 70\% & 0.34s \\
\bottomrule
\end{tabular}
}
\vspace{-1.5em}
\end{table}

\noindent\textbf{Slow-Fast Denoising Ratio.}
Finally, we evaluate the computation allocation between the Dreamer and the Refiner. The 3:7 slow--fast configuration achieves the best overall result, suggesting that LiMA requires sufficient computation for the Dreamer to generate informative long-term intent, while allocating more capacity to the Refiner for high-frequency and fine-grained motion correction. Overall, these ablation results demonstrate that LiMA's robust dexterous manipulation performance depends not only on the dual-system architecture itself, but also on an appropriate computation allocation between long-term imagination and real-time control.

\subsection{Generalization}

\noindent
\rlap{%
  \makebox[\linewidth][r]{%
    \raisebox{0pt}[0pt][0pt]{%
      \begin{minipage}[t]{0.35\linewidth}
        \centering
        \vspace{-10pt}
        \includegraphics[width=\linewidth]{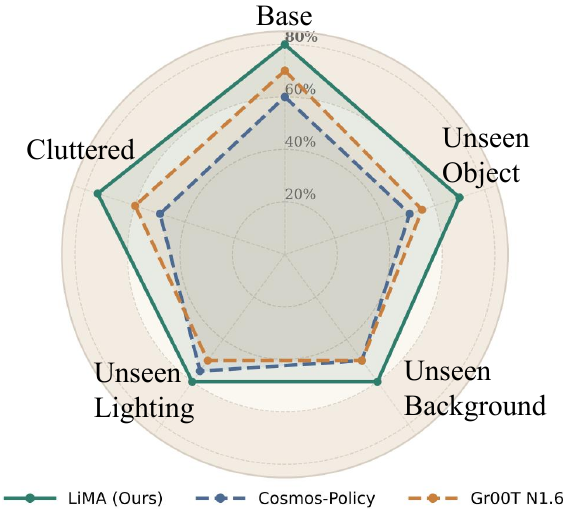}
        \vspace{-6pt}
        \captionof{figure}{\textbf{Generalizations.}}
        \label{fig:generalization}
      \end{minipage}%
    }%
  }%
}%
\hangindent=-0.36\linewidth
\hangafter=-13
Beyond standard task evaluation, we further assess LiMA's out-of-distribution (OOD) generalization on the Cook Rice task, which requires coordinated bimanual execution across multiple task stages. We introduce four common visual perturbations: unseen backgrounds, novel object instances, unseen illumination conditions, and cluttered scenes. These settings evaluate different aspects of visual robustness. For each setting, we conduct 20 real-world trials, and Gr00T N1.6 and Cosmos-Policy are evaluated under the same protocol. As shown in \Cref{fig:generalization}, LiMA consistently achieves higher success rates across all OOD settings. Its performance under unseen backgrounds and illumination conditions indicates that the policy does not merely memorize environment-specific textures or appearance statistics. LiMA also remains robust in cluttered scenes, suggesting that the model can preserve task-relevant intent despite additional visual distractors. Most notably, LiMA maintains a 70\% success rate when manipulating novel object instances, demonstrating that it learns transferable interaction patterns rather than relying exclusively on object-specific visual cues observed during training. We attribute this robustness to the complementary roles of the two systems. The Future Dreamer captures long-horizon task structure and global object relationships, providing a relatively stable intent prior under visual changes. Meanwhile, the Motion Refiner continuously integrates the latest observations to correct local execution and accommodate deviations between imagined futures and the current physical state. This separation between task-level intent generation and observation-conditioned motion refinement reduces sensitivity to individual sources of visual variation. 
\par

\section{Limitations}
While LiMA excels in diverse dexterous tasks, its performance degrades under severe visual occlusion and low visual variance, where vision-based planning is inherently constrained. The current reliance on camera inputs limits the Dreamer’s ability to maintain accurate trajectories when key features are obscured. We believe integrating tactile feedback is a crucial next step; incorporating haptic sensing would compensate for visual uncertainty and enable the Refiner to manage contact-rich interactions more robustly when visual information is compromised or temporarily unreliable. In addition, the reported inference latency is measured using a general implementation without deployment-specific acceleration, such as operator fusion, model compilation, quantization, or multi-GPU parallelism. Although this setting enables a consistent architecture-level comparison across different methods, it may not reflect the maximum achievable efficiency of LiMA. Future work will investigate fused computational kernels, reduced-precision inference, parallel execution of the Dreamer and Refiner, and hardware-aware multi-GPU deployment to further reduce latency and improve real-time performance.

\section{Conclusion}
\label{sec:conclusion}
In this paper, we presented \textbf{LiMA}, an asynchronous diffusion-based framework for bridging long-term imagination and high-frequency reactive execution. LiMA coordinates the Future Dreamer and the Motion Refiner through a dual-system architecture, enabling long-horizon foresight while preserving real-time responsiveness. A Latent Schrödinger Bridge Coupling mechanism further aligns strategic intent with clean execution latents. Extensive real-world experiments show that LiMA improves complex long-horizon manipulation while maintaining robustness and efficient closed-loop control.

%===============================================================================

\clearpage
% The acknowledgments are automatically included only in the final and preprint versions of the paper.
\acknowledgments{This work was supported by the National Natural Science Foundation of China (62476011), the Beijing Natural Science Foundation (L252060). We are grateful to Xiansheng Chen for his support with the hardware setup, and to Yuqi Cao, Liangwei Hu, and Liwen Tong for their valuable insights and in-depth discussions on the mathematical derivations.}

%===============================================================================

\bibliography{example}  % .bib

\clearpage
 
\appendix
\noindent{\large\bfseries Appendix}
\section{Additional Method Details}

\subsection{\textbf{Additional Preliminaries}}
\label{app:additional_preliminaries}

\noindent\textbf{Latent Schr\"odinger Bridge Coupling.}
Schr\"odinger Bridge formulates a stochastic transport problem between two prescribed boundary distributions. 
Unlike standard diffusion generation, which typically starts from an unstructured Gaussian prior during sampling, Schr\"odinger Bridge constructs a bridge process whose endpoints are constrained by two informative boundary distributions. 
I$^2$SB~\cite{liu2023i2sb} provides a practical instantiation of this idea by defining an analytic posterior over intermediate bridge states between a structured source endpoint and a clean target endpoint.

Given a paired boundary sample $(X_0, X_1)$, the intermediate state $X_t$ is sampled from:
\begin{equation}
q(X_t \mid X_0, X_1)
=
\mathcal{N}
\left(
X_t;
\mu_t(X_0,X_1),
\Sigma_t \mathbf{I}
\right),
\end{equation}
where
\begin{equation}
\mu_t(X_0,X_1)
=
\frac{\bar{\sigma}_t^2}
{\sigma_t^2+\bar{\sigma}_t^2} X_0
+
\frac{\sigma_t^2}
{\sigma_t^2+\bar{\sigma}_t^2} X_1,
\qquad
\Sigma_t
=
\frac{\sigma_t^2\bar{\sigma}_t^2}
{\sigma_t^2+\bar{\sigma}_t^2}.
\end{equation}
Here, $\sigma_t^2=\int_0^t \beta_{\tau}d\tau$ and 
$\bar{\sigma}_t^2=\int_t^1 \beta_{\tau}d\tau$ denote the accumulated variances along the bridge. 
The mean path $\mu_t(X_0,X_1)$ defines the deterministic interpolation between the two endpoints, while $\Sigma_t \mathbf{I}$ introduces controlled stochasticity around this path. 
This gives a boundary-conditioned posterior rather than an unconditional Gaussian prior.

The endpoint behavior of this posterior clarifies the bridge direction. 
At $t=0$, we have $\sigma_t^2=0$, which makes $\mu_t(X_0,X_1)=X_0$ and $\Sigma_t=0$. 
At $t=1$, we have $\bar{\sigma}_t^2=0$, which makes $\mu_t(X_0,X_1)=X_1$ and $\Sigma_t=0$. 
Equivalently, a sampled intermediate state can be written as 
$\mu_t(X_0,X_1)+\sqrt{\Sigma_t}\epsilon$, where $\epsilon\sim\mathcal{N}(0,\mathbf{I})$. 
Thus, the stochastic term vanishes at both endpoints, while intermediate timesteps retain controlled stochasticity between the two meaningful latent endpoints. 
This differs from standard diffusion generation, where the sampling trajectory is typically anchored by an uninformative Gaussian prior.

In LiMA, we use this boundary-conditioned posterior as the mathematical interface between the Future Dreamer and the Motion Refiner. 
The key adaptation lies in redefining the two bridge boundaries for asynchronous robotic control: the structured source boundary $X_1$ is instantiated as the Dreamer's strategic intent prior,
\begin{equation}
X_1 \equiv \mathcal{Z}_{\text{int}},
\end{equation}
while the clean target boundary $X_0$ is instantiated as the Refiner's clean joint execution latent,
\begin{equation}
X_0 \equiv \mathcal{Z}_{\text{out}}^{(0)}.
\end{equation}
Substituting these two boundaries into the I$^2$SB posterior yields the latent bridge used in LiMA:
\begin{equation}
q
\left(
\mathcal{Z}_{\text{out}}^{(t)}
\mid
\mathcal{Z}_{\text{out}}^{(0)},
\mathcal{Z}_{\text{int}}
\right)
=
\mathcal{N}
\left(
\mathcal{Z}_{\text{out}}^{(t)};
\mu_t
\left(
\mathcal{Z}_{\text{out}}^{(0)},
\mathcal{Z}_{\text{int}}
\right),
\Sigma_t \mathbf{I}
\right).
\end{equation}
This turns the original boundary-to-boundary bridge into an intent-to-execution latent coupling mechanism. 
As a result, LiMA initializes refinement from a structured long-horizon intent prior rather than an unconditional Gaussian latent, while retaining stochastic refinement capacity through the bridge posterior.

\subsection{\textbf{Additional Implementation Details}}
\label{app:implementation_details}

\noindent\textbf{Latent token layout.}
Following the latent tensor convention of Cosmos-Policy~\cite{kim2026cosmos}, LiMA represents multimodal information in a five-dimensional latent tensor. 
Each latent tensor is organized as
\begin{equation}
\mathcal{Z} \in 
\mathbb{R}^{B \times C \times N_{\mathrm{slot}} \times H \times W},
\end{equation}
where $B$ is the batch size, $C=16$ is the latent channel dimension, $N_{\mathrm{slot}}=8$ denotes the number of organized latent slots, and $H=W=28$ are the spatial latent resolutions. 
This slot-based organization provides an explicit interface between visual observations, action priors, and future imagination tokens.

The Future Dreamer and Motion Refiner use compatible slot layouts, each with $N_{\mathrm{slot}}=8$. 
Specifically, the Future Dreamer uses the slot set $\mathcal{S}_{\mathrm{dream}}$ = 
[\texttt{blank}, \texttt{left\_wrist}, \texttt{right\_wrist}, \texttt{primary}, \texttt{action}, \texttt{left\_future}, \texttt{right\_future}, \texttt{primary\_future}]. 
The Motion Refiner uses the slot set $\mathcal{S}_{\mathrm{refine}}$ = 
[\texttt{blank}, \texttt{left\_wrist}, \texttt{right\_wrist}, \texttt{primary}, \texttt{arm\_action}, \texttt{left\_hand\_action}, \texttt{right\_hand\_action}, \texttt{primary\_future}]. 
This compatible slot organization allows the Dreamer's intent latent $\mathcal{Z}_{\text{int}}$ to be directly consumed by the Motion Refiner as a structured prior boundary. 
Therefore, the latent Schr\"odinger bridge can be constructed in the shared latent space without introducing an additional projection or alignment network.

\noindent\textbf{Training hyperparameters.}
For training LiMA, the Future Dreamer is initialized from the pretrained Cosmos-Predict2-2B checkpoint, while the Motion Refiner is randomly initialized and trained from scratch. We use a batch size of 14 and optimize the model with a peak learning rate of $1.0 \times 10^{-4}$. The learning rate is warmed up for 2.5K steps, then decayed to $0.3$ of the peak value within the first 40K steps. After that, the learning rate is kept constant at $0.06$ of the peak learning rate.

\section{Additional Experimental Setup}

\begin{figure}
    \centering
    \captionsetup{type=figure}
    \includegraphics[width=0.7\textwidth]{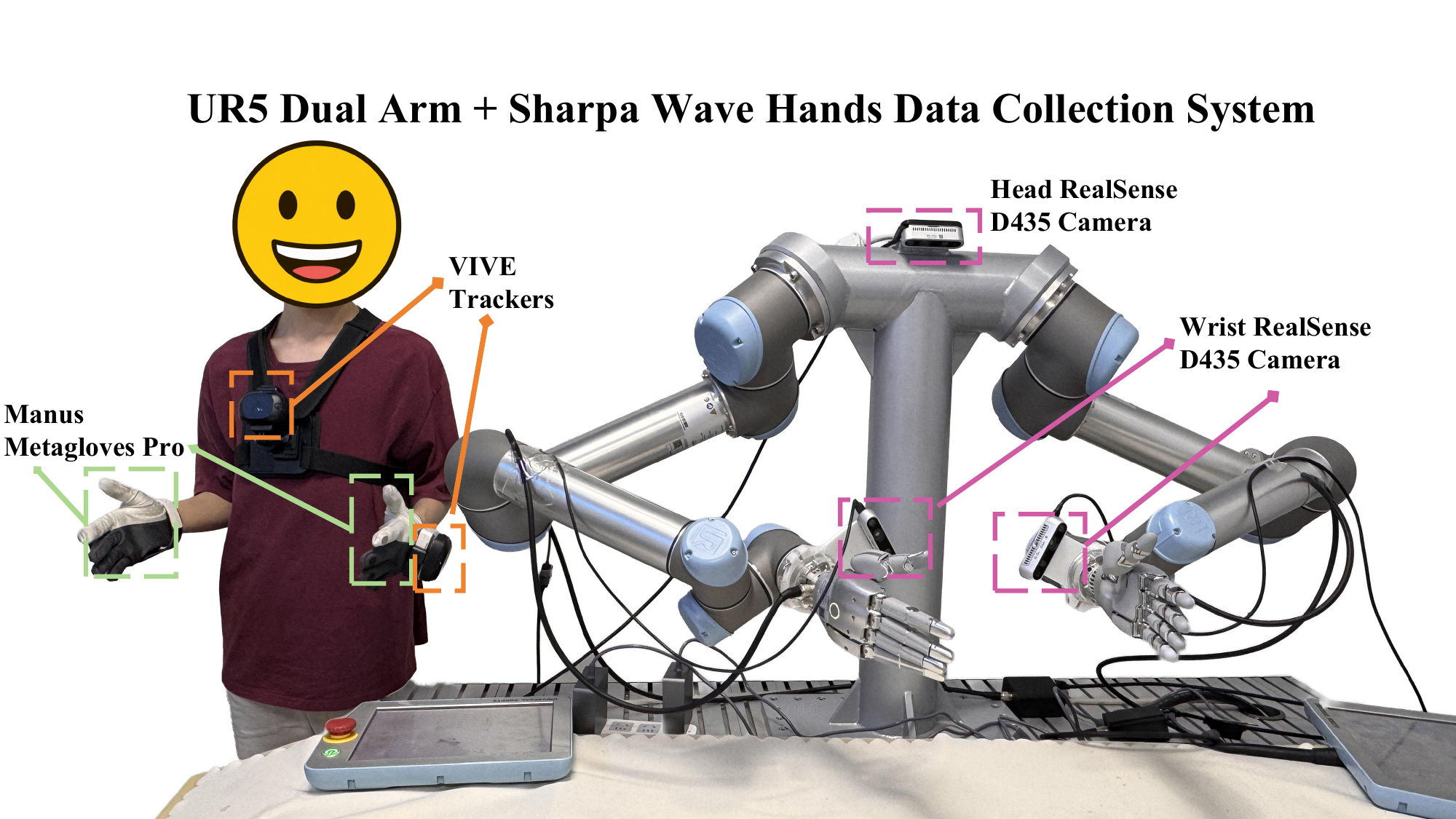}
    \caption{
    Overview of the real-world bimanual manipulation platform.
    The system consists of two UR5 robotic arms equipped with SharpaWave dexterous hands, a three-view RealSense D435 camera setup with one head-mounted camera and two wrist-mounted cameras, and a wearable teleoperation interface based on VIVE Trackers and MetaGlove Pro gloves for expert demonstration collection.
    }
    \label{fig:robot_system}
    \vspace{-1em}
\end{figure}

\subsection{\textbf{Robot System Setup}}
\label{app:robot_system_setup}

We further provide implementation details of the real-world robot system, including the system operation pipeline, demonstration collection, and policy execution protocol. 
The experimental platform follows the bimanual dexterous manipulation setup described in the main paper, as illustrated in Figure~\ref{fig:robot_system}. 
It consists of two UR5 robotic arms, two SharpaWave five-fingered dexterous hands, three Intel RealSense D435 cameras, VIVE Trackers, and MetaGlove Pro gloves.

\noindent\textbf{Perception System.}
The three RealSense D435 cameras provide one global view and two local wrist views. 
The head-mounted camera captures the overall task scene, including target objects, the manipulation workspace, and task progress, while the two wrist-mounted cameras are installed near the end-effectors of the left and right robotic arms to provide local visual observations around the contact regions. 
All cameras are fixed before the experiments to ensure consistent multi-view observations within the same robot workspace.

\noindent\textbf{Demonstration Collection.}
During human demonstration collection, the operator uses VIVE Trackers to record the relative spatial trajectories of both wrists, which are transformed and mapped to end-effector reference trajectories for the two UR5 arms. 
Meanwhile, MetaGlove Pro gloves record the finger joint configurations of both hands, which are mapped to the 22-DoF SharpaWave dexterous hands through a joint-space kinematic retargeting pipeline. 
This process produces synchronized dual-arm trajectories, dual-hand joint trajectories, multi-view visual observations, and language task instructions.

\noindent\textbf{Policy Execution.}
During policy execution, each policy generates a short-horizon action chunk based on the current observations. 
The predicted action chunk contains motion commands for the two UR5 arms and joint commands for the left and right SharpaWave dexterous hands. 
Before each real-world trial, the experimental scene is reset to a predefined initial state, including the robot initial posture and the corresponding task instruction. 
All methods are evaluated under the same robot platform, camera configuration, task initialization protocol, and success criteria, ensuring that the comparison between LiMA and the baselines mainly reflects differences in policy design rather than hardware conditions or evaluation settings.

\subsection{\textbf{Baseline Settings}}
\label{app:baseline_settings}

We compare LiMA with representative baselines covering integrated world-action modeling, vision-language-action modeling, and video-prediction-based policy learning. 
All baselines are evaluated under the same real-world robot platform, camera configuration, task initialization protocol, and success criteria as LiMA. 
For each baseline, we follow the training or fine-tuning protocol recommended by its original paper whenever applicable, and adapt its action outputs to the same bimanual dexterous robot execution interface.

\noindent\textbf{Cosmos-Policy.}
Cosmos-Policy~\cite{kim2026cosmos} is used as an integrated world-action modeling baseline. 
It maps robot actions into the visual latent space and jointly denoises action latents and image latents within a unified DiT-based denoising process. 
This makes Cosmos-Policy the closest integrated generative baseline to LiMA, as both methods model future visual information and robot actions within a shared generative framework. 

\noindent\textbf{GR00T N1.6.}
GR00T N1.6~\cite{gr00tn1_2025} is used as a representative state-of-the-art vision-language-action baseline. 
It takes robot visual observations and language instructions as inputs to a vision-language model for semantic understanding, and conditions an action head with the robot state to generate continuous actions through a denoising process. 
In our setting, we found that the officially recommended 2,000 training steps did not fully expose its best performance on our bimanual dexterous manipulation tasks. 
Therefore, we report the results after training GR00T N1.6 for 50,000 steps on 1 NVIDIA H100 GPU.

\noindent\textbf{VPP.}
VPP~\cite{hu2024video} is used as an inverse-dynamics-model baseline. 
It uses the future visual predictions produced by a video model as conditioning information and feeds them into an action DiT to generate robot actions. 
This baseline represents the paradigm of first predicting future visual states and then deriving actions from the predicted visual condition. 

\noindent\textbf{InternVLA-A1.}
InternVLA-A1~\cite{internvla_a1} adopts a Mixture-of-Transformers (MoT) architecture that unifies scene understanding, generative foresight, and action modeling within a single framework, enabling the model not only to acquire a rich representation of the environment but also to leverage predicted future visual observations to guide action generation. To accommodate the full joint configuration of our dual-arm, dual-hand system—comprising two 6-DoF robot arms and two 22-DoF dexterous hands (a total of 56‑DoF)—we extend the action dimensionality accordingly and initialize the model from the pretrained InternVLA-A1-3B checkpoint. 

To ensure a fair comparison, all baselines are trained or fine-tuned on the same real-world demonstration dataset whenever task-specific training is required. 
They use the same task instructions, multi-view robot observations, robot state inputs, and evaluation protocol as LiMA. 
Since different baselines adopt different native action representations, we adapt their output interfaces to the same bimanual execution space, including arm motion commands and dexterous hand joint commands.

\section{Additional Evaluation Details}

\begin{figure}
    \centering
    \captionsetup{type=figure}
    \includegraphics[width=\textwidth]{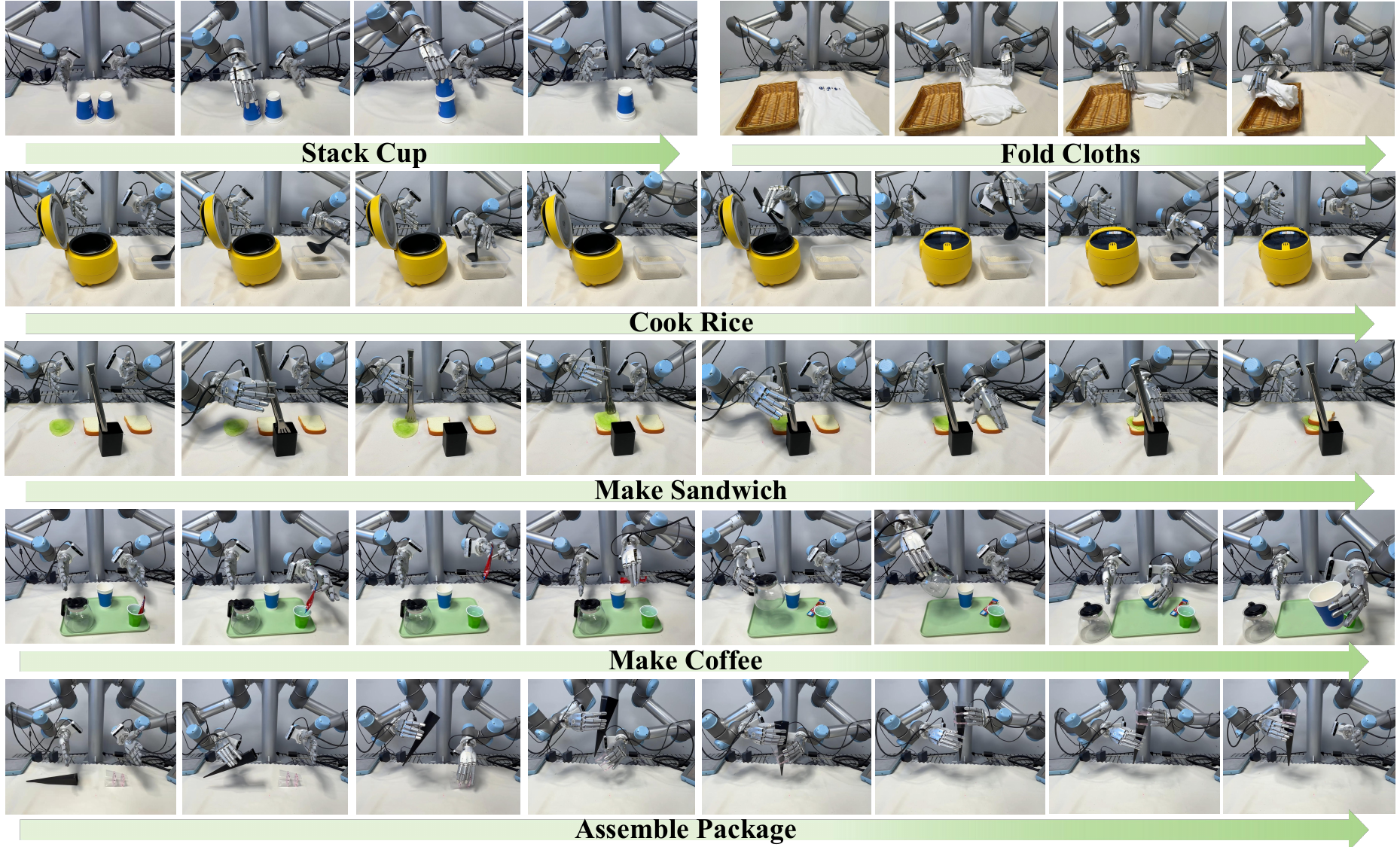}
    \caption{\textbf{Visualization of task progress.}}
    \label{fig:task_progress}
\end{figure}

\subsection{\textbf{Task Details}}
\label{app:task_details}

We provide additional details on the real-world manipulation tasks used in our evaluation. 
We design six dexterous manipulation tasks with varying temporal horizons and interaction complexity, including Stack Cup, Roll T-shirt, Cook Rice, Make Sandwich, Make Coffee, and Assemble Package. 
These tasks cover both short-horizon single-arm manipulation and long-horizon bimanual coordination, requiring the robot to generate continuous actions from multi-view visual observations and language instructions. Visualizations of these real-world tasks are shown in Figure~\ref{fig:task_progress}.

\noindent\textbf{Stack Cup.}
Stack Cup is a short-horizon single-arm manipulation task. 
The robot is required to use its right hand to gently grasp a cup, align it above another cup, and place it down to complete the stacking process. 
This task mainly evaluates basic grasping, alignment, and precise placement ability.

\noindent\textbf{Roll T-shirt.}
Roll T-shirt is a bimanual deformable-object manipulation task. 
The robot first uses both hands to grasp two corners of the T-shirt and folds it forward for several steps until the garment becomes easier to grasp with one hand. 
Finally, the robot uses its right hand to pick up the folded garment and place it into a basket. 
This task evaluates the model's ability to localize manipulation regions under diverse cloth deformations and to recognize different task stages during deformable-object manipulation.

\noindent\textbf{Cook Rice.}
Cook Rice is a long-horizon multi-stage task. 
The robot first uses its left hand to gently grasp a spoon, scoop rice, and steadily transfer the rice into a rice cooker. 
After pouring the rice, the robot uses its right hand to close the rice cooker and then returns the spoon with its left hand. 
This task requires smooth motion over a longer horizon and consistent maintenance of task-level intent across multiple manipulation stages.

\noindent\textbf{Make Sandwich.}
Make Sandwich is a long-horizon tool-use task. 
The robot first uses its right hand to pick up a pair of tongs, grasp a piece of lettuce, and place it onto a slice of bread. 
After accurately returning the tongs, the robot uses its left hand to pick up another slice of bread and place it on top of the lettuce. 
This task mainly evaluates tool use, fine-grained object interaction, and precise multi-step placement.

\noindent\textbf{Make Coffee.}
Make Coffee is a long-horizon multi-object interaction task. 
The robot first uses its left hand to pick up a coffee bag and pour coffee into a cup, then uses its right hand to pick up a kettle and pour water into the cup. 
Finally, the robot uses its left hand to grasp the cup and hand it to a person in front of the workspace. 
For safety and hardware protection, including water and dust resistance considerations, this task is performed with simulated pouring actions rather than real liquid transfer. 
The task mainly evaluates the model's ability to handle multi-object interactions over a long temporal horizon.

\noindent\textbf{Assemble Package.}
Assemble Package is a long-horizon bimanual coordination task. 
The robot first uses its right hand to pick up a black package component and then uses its left hand to pick up a transparent package component. 
The two hands then cooperate to align and combine the two components into a complete package. 
This task evaluates bimanual coordination, spatial alignment, and execution stability in a structured assembly scenario.

\begin{table*}[t]
\centering
\small
\caption{\textbf{Progress stage definitions for real-world tasks.} 
Each task is decomposed into several progress stages for computing Progress Success Rate (PSR).}
\label{tab:progress_stage_definitions}
\begin{tabularx}{\textwidth}{lXXXXX}
\toprule
\textbf{Task} & \textbf{P1} & \textbf{P2} & \textbf{P3} & \textbf{P4} & \textbf{P5} \\
\midrule
Stack Cup 
& Grasp cup 
& Stack cup 
& -- 
& -- 
& -- \\

Roll T-shirt 
& Grasp corners 
& Fold garment 
& Pick and place 
& -- 
& -- \\

Cook Rice 
& Grasp spoon 
& Scoop rice 
& Transfer rice 
& Close cooker 
& Return spoon \\

Make Sandwich 
& Pick up tongs 
& Grasp lettuce 
& Place lettuce and return tongs 
& Cover with bread
& -- \\

Make Coffee 
& Pick up coffee bag 
& Pick up kettle 
& Hand over cup
& --
& -- \\

Assemble Package 
& Pick up black part 
& Pick up transparent part 
& Combine parts
& --
& -- \\
\bottomrule
\end{tabularx}
\end{table*}

\begin{figure}[b]
    \centering
    \captionsetup{type=figure}
    \includegraphics[width=\textwidth]{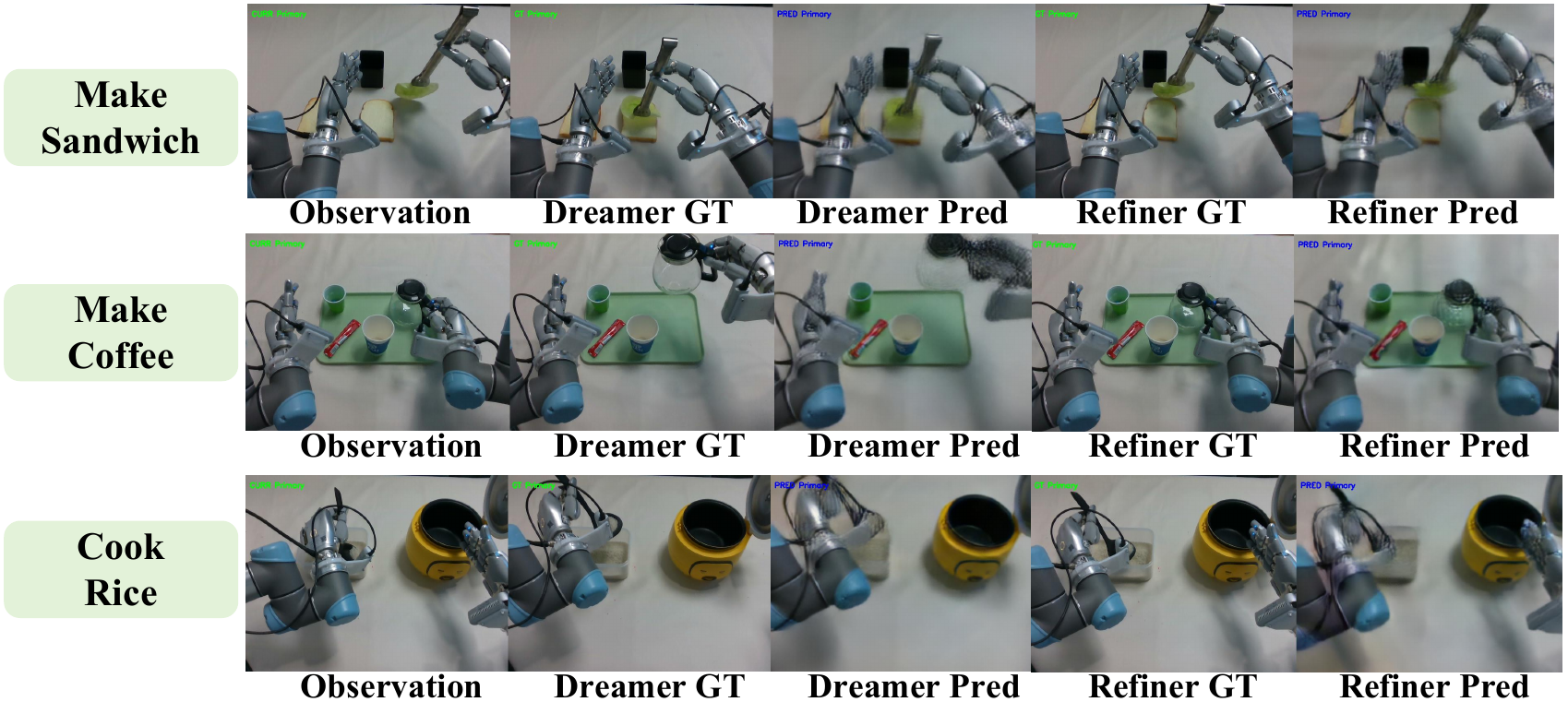}
    \caption{\textbf{Visualization of Dreamer \& Refiner Generations.}}
    \label{fig:Generations}
\end{figure}

\begin{table*}[t]
\centering
\small
\caption{\textbf{Per-stage success rates for progress evaluation.} 
We report the success rate of each progress stage to analyze where different methods fail during task execution.}
\label{tab:per_stage_success_rates}
\begin{tabular}{llcccccc}
\toprule
\textbf{Task} & \textbf{Method} & \textbf{P1(\%)} & \textbf{P2(\%)} & \textbf{P3(\%)} & \textbf{P4(\%)} & \textbf{P5(\%)} & \textbf{PSR(\%)} \\
\midrule

Stack Cup 
& LiMA & \textbf{100} & \textbf{90.0} & -- & -- & -- & \textbf{95.0} \\
& Cosmos-Policy & 85.0 & 80.0 & -- & -- & -- & 82.5 \\
& GR00T N1.6 & 90.0 & 85.0 & -- & -- & -- & 87.5 \\
& VPP & 80.0 & 80.0 & -- & -- & -- & 80.0 \\
& InternVLA-A1 & 90.0 & \textbf{90.0} & -- & -- & -- & 90.0 \\
\midrule

Roll T-shirt 
& LiMA & 95.0 & 80.0 & \textbf{75.0} & -- & -- & 83.3 \\
& Cosmos-Policy & 80.0 & 70.0 & 60.0 & -- & -- & 70.0 \\
& GR00T N1.6 & 75.0 & 70.0 & 70.0 & -- & -- & 71.7 \\
& VPP & 60.0 & 55.0 & 55.0 & -- & -- & 56.7 \\
& InternVLA-A1 & \textbf{100} & \textbf{90.0} & \textbf{75.0} & -- & -- & \textbf{88.3} \\
\midrule

Cook Rice 
& LiMA & \textbf{85.0} & \textbf{85.0} & \textbf{80.0} & \textbf{80.0} & \textbf{80.0} & \textbf{82.0} \\
& Cosmos-Policy & 75.0 & 70.0 & 70.0 & 60.0 & 60.0 & 67.0 \\
& GR00T N1.6 & 80.0 & 70.0 & 70.0 & 70.0 & 70.0 & 72.0 \\
& VPP & 65.0 & 60.0 & 60.0 & 60.0 & 55.0 & 60.0 \\
& InternVLA-A1 & \textbf{85.0} & 80.0 & \textbf{80.0} & 75.0 & 70.0 & 78.0 \\
\midrule

Make Sandwich 
& LiMA & 90.0 & \textbf{85.0} & \textbf{75.0} & \textbf{70.0} & -- & \textbf{80.0} \\
& Cosmos-Policy & 90.0 & 70.0 & 65.0 & 60.0 & -- & 71.3 \\
& GR00T N1.6 & 80.0 & 75.0 & 70.0 & 65.0 & -- & 72.5 \\
& VPP & 75.0 & 60.0 & 60.0 & 50.0 & -- & 61.3 \\
& InternVLA-A1 & \textbf{100} & 75.0 & 70.0 & 65.0 & -- & 77.5 \\
\midrule

Make Coffee 
& LiMA & \textbf{80.0} & 70.0 & \textbf{60.0} & -- & -- & 70.0 \\
& Cosmos-Policy & 70.0 & 65.0 & 55.0 & -- & -- & 63.3 \\
& GR00T N1.6 & \textbf{80.0} & \textbf{80.0} & 55.0 & -- & -- & \textbf{71.7} \\
& VPP & 60.0 & 60.0 & 50.0 & -- & -- & 56.7 \\
& InternVLA-A1 & \textbf{80.0} & 70.0 & 50.0 & -- & -- & 66.7 \\
\midrule

Assemble Package 
& LiMA & \textbf{70.0} & \textbf{70.0} & \textbf{50.0} & -- & -- & \textbf{63.3} \\
& Cosmos-Policy & 60.0 & 50.0 & 45.0 & -- & -- & 51.7 \\
& GR00T N1.6 & \textbf{70.0} & 65.0 & \textbf{50.0} & -- & -- & 61.7 \\
& VPP & 60.0 & 45.0 & 40.0 & -- & -- & 48.3 \\
& InternVLA-A1 & 65.0 & 60.0 & \textbf{50.0} & -- & -- & 58.3 \\

\bottomrule
\end{tabular}
\end{table*}

\subsection{\textbf{Evaluation Metrics}}
\label{app:evaluation_metrics}

We provide additional details on the evaluation metrics used in our real-world robot experiments. 
We mainly evaluate different methods using Success Rate (SR), Progress Success Rate (PSR), and inference time. 
SR measures whether a policy can complete the entire task. 
For each real-world trial, the trial is considered successful only if the robot completes all predefined task goals in the correct order. 
Thus, SR reflects the final task-level success of each policy.

For long-horizon multi-stage tasks, binary SR alone cannot fully reflect partial task completion. 
Therefore, we also use the Progress Success Rate (PSR) to measure the progress of the task. 
The PSR is calculated on the basis of the proportion of progress stages successfully completed in each trial. 
The progress stages used for the PSR calculation are defined in Table~\ref{tab:progress_stage_definitions}, and the success rates per-stage are reported in Table~\ref{tab:per_stage_success_rates}. 
This enables a more fine-grained analysis of where different methods fail during task execution.
In addition to these quantitative progress metrics, we visualize LiMA's predicted ego-centric future observations and the corresponding ground-truth observations for three long-horizon tasks in Figure~\ref{fig:Generations}.

We also report inference time to evaluate real-time deployability in closed-loop robot control. 
Inference time denotes the average time required for a policy to generate one executable action chunk from the current observation. 
For synchronous inference models, this time is measured by the forward-pass time of each complete policy inference. 
For LiMA, since the Future Dreamer is updated at a lower frequency while the Motion Refiner runs at the high-frequency control loop, we report the effective average inference time under asynchronous execution. 
Specifically, it is computed as the sum of one Future Dreamer inference time and all Motion Refiner inference times within one asynchronous update cycle, divided by the number of action chunks generated in that cycle. 
All inference-time measurements are collected under the same robot platform and execution protocol to ensure a fair comparison.

\subsection{\textbf{Generalization Evaluation}}
\label{app:generalization_evaluation}

\begin{figure}
    \centering
    \captionsetup{type=figure}
    \includegraphics[width=\textwidth]{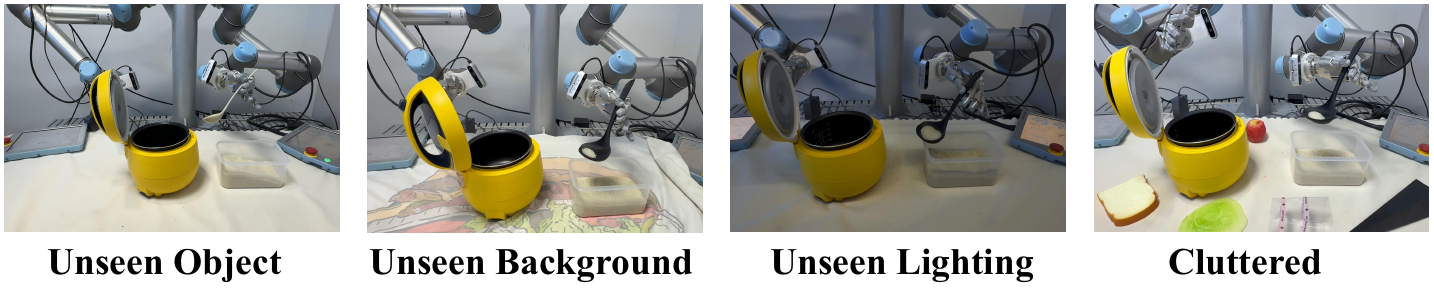}
    \caption{\textbf{Visualization of Generalizations.}}
    \label{fig:Generalizations}
\end{figure}

We provide additional qualitative analysis for the generalization evaluation, with task visualizations shown in Figure~\ref{fig:Generalizations}. 
The quantitative success rates are reported in the main paper. 
In the unseen-object setting, LiMA still maintains a success rate of 70\%, indicating that the policy can adapt to changes in task-relevant object appearance while preserving the overall manipulation intent. 
In the cluttered-scene setting, LiMA achieves a success rate of 75\%, suggesting that the proposed architecture can effectively focus on task-relevant regions and remain robust to irrelevant visual distractions. Compared with LiMA, Cosmos-Policy shows more frequent grasping failures under these perturbed scenes. This suggests that a unified DiT-based world-action model may be more sensitive to visual clutter and local object appearance shifts during contact-rich execution. GR00T N1.6 also exhibits failure modes that are not observed under the base setting, such as skipping the rice-scooping stage in the Cook Rice task.

For unseen-background and unseen-lighting settings, LiMA shows larger success-rate fluctuations than in the base setting, as these perturbations directly affect the visual appearance of the task region. 
Nevertheless, LiMA still outperforms the baselines under these conditions. 
These results suggest that the asynchronous Dreamer--Refiner design and latent intent-to-execution coupling help LiMA maintain stronger robustness when both task-relevant and task-irrelevant visual factors are perturbed.

\section{\textbf{Failure Cases}}
\label{app:failure_cases}

We further analyze the typical failure cases of LiMA in real-world deployment. 
The failures mainly come from two aspects. 
First, LiMA is sensitive to the field of view of the installed cameras. 
The policy relies on multi-view visual observations to maintain long-horizon task intent and local motion correction, so the task scene needs to be sufficiently covered by the cameras. 
For example, in the Assemble Package task, some manipulation steps may partially move outside the camera view due to the limited visual coverage. 
In addition, transparent objects are harder to perceive reliably from RGB observations, which further limits LiMA's long-horizon planning and execution stability. 
This partially explains why the success rate of Assemble Package is lower than that of other tasks.

Second, LiMA is still limited in perceiving fine-grained contact information from vision alone. 
For example, in the Roll T-shirt task, when the manipulated garment has a color similar to the surrounding environment, the model may be less sensitive to the exact contact region. 
In some trials, the robot occasionally pinches empty space instead of the garment, leading to failed folding or grasping. 
These results suggest that although LiMA can maintain stable task-level intent in most cases, its performance is still constrained by visual observability and contact-state estimation. 
In future work, these limitations can be mitigated by using cameras with a wider field of view and incorporating tactile sensing to provide more reliable contact feedback.

\end{document}